\documentclass[10pt,twocolumn,letterpaper,dvipdfmx]{article}

\usepackage{iccv}
\usepackage{times}
\usepackage{epsfig}
\usepackage{graphicx}
\usepackage{amsmath}
\usepackage{amssymb}

\usepackage[subrefformat=parens]{subcaption}
\usepackage{multirow}
\usepackage{bm}

\usepackage[breaklinks=true,bookmarks=false]{hyperref}

\iccvfinalcopy 

\def\iccvPaperID{****} 

\ificcvfinal\fi

\begin{document}

\title{Deep Ensemble Collaborative Learning by using Knowledge-transfer Graph \\
for Fine-grained Object Classification}

\author{
    Naoki Okamoto, Soma Minami, Tsubasa Hirakawa,\\
    Takayoshi Yamashita, Hironobu Fujiyoshi\\
    Chubu University\\
    {\tt \small \{naok@mprg.cs, minami@mprg.cs, hirakawa@mprg.cs, takayoshi@isc, fujiyoshi@isc\}.chubu.ac.jp}\\
}

\maketitle
\ificcvfinal\thispagestyle{empty}\fi

\begin{abstract}
Mutual learning, in which multiple networks learn by sharing their knowledge, improves the performance of each network.
However, the performance of ensembles of networks that have undergone mutual learning does not improve significantly from that of normal ensembles without mutual learning, even though the performance of each network has improved significantly.
This may be due to the relationship between the knowledge in mutual learning and the individuality of the networks in the ensemble.
In this study, we propose an ensemble method using knowledge transfer to improve the accuracy of ensembles by introducing a loss design that promotes diversity among networks in mutual learning.
We use an attention map as knowledge, which represents the probability distribution and information in the middle layer of a network.
There are many ways to combine networks and loss designs for knowledge transfer methods.
Therefore, we use the automatic optimization of knowledge-transfer graphs to consider a variety of knowledge-transfer methods by graphically representing conventional mutual-learning and distillation methods and optimizing each element through hyperparameter search.
The proposed method consists of a mechanism for constructing an ensemble in a knowledge-transfer graph, attention loss, and a loss design that promotes diversity among networks.
We explore optimal ensemble learning by optimizing a knowledge-transfer graph to maximize ensemble accuracy.
From exploration of graphs and evaluation experiments using the datasets of Stanford Dogs, Stanford Cars, and CUB-200-2011, we confirm that the proposed method is more accurate than a conventional ensemble method.
\end{abstract}

\section{Introduction}
\begin{figure}[t]
	\centering
	\includegraphics[width=8.3cm]{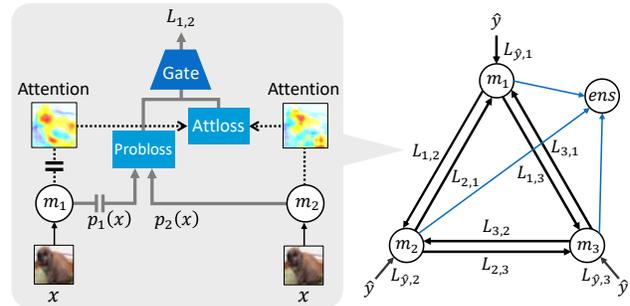}
	\caption{Knowledge-transfer graph (for 3 nodes) introducing mechanism for constructing an ensemble and attention loss. Loss calculation shows knowledge transfer from $m_1$ to $m_2$.}
	\label{fig:1}
	\vspace{-1mm}
\end{figure}

When networks are trained under the same conditions, such as network architecture and data set, they produce different errors, even though they have the same level of accuracy, depending on random factors such as the initial values of the network and data selected as mini-batches.
This change can be said to indicate that the network acquires different knowledge during training depending on the training conditions.
ensemble and knowledge-transfer methods have performed well on a variety of problems by using multiple networks with different weight parameters for training and inference.

An ensemble method executes inference using multiple learned networks.
The inference is done by averaging the output of each network for the input samples. This improves ensemble accuracy compared with inference using a single network.
Due to the nature of using multiple networks, it is also effective in various problem settings such as adversarial attack and out-of-distribution detection \cite{semi-supEns,FewShotEns,AdversarialEns,OutOfDisRns}.
Compared with a single network, the computational cost increases with the number of networks used in an ensemble, so an efficient ensemble methods have been proposed \cite{BatchEns,HypEns,CL,ONE}.

Knowledge transfer is a learning method in which a network shares the knowledge acquired in learning with the goal of network compression and network performance improvement.
There are two types of knowledge-transfer methods: unidirectional \cite{KD} and bidirectional \cite{DML}.
Unidirectional knowledge transfer \cite{KD} is a one-way knowledge-transfer method that learns an untrained network with shallow layers using a learned network with deep layers.
Bidirectional knowledge transfer \cite{DML} is a mutual-learning method using multiple untrained networks.
These two types of methods use probability distributions as knowledge.
Knowledge-transfer methods have also been proposed for various conditions such as network size and knowledge to be used, e.g., knowledge transfer to unlearned networks of similar layer depths and knowledge transfer using features of intermediate layers as network knowledge \cite{FitNets,IRG,FSP_matrix,similarity-preserving,correlation_congruence,VID,Learning_Metrics_from_Teachers,CRD,AT,Grad-CAM++}. 
With these methods, the combination of networks and the direction of knowledge transfer are manually designed.
A knowledge-transfer graph is used of considering various knowledge-transfer methods by optimizing the knowledge-transfer direction and network combination as hyperparameters.
In knowledge-transfer graphs \cite{KTG}, networks are represented as nodes and knowledge propagation from node to node as edges, and knowledge transfer is represented as a directed graph.
Since there is no need to fix the network architecture or knowledge-transfer method in advance, various learning methods can be represented by combining them.

In this study, we propose an ensemble method using knowledge transfer to improve ensemble accuracy by promoting diversity among networks during training.
We focus on mutual learning and knowledge-transfer graphs for this purpose.
In addition to traditional knowledge transfer, we use mutual learning to separate the outputs between networks.
If the probability distributions are directly separated, the performance of the networks may degrade.
Therefore, we promote diversity from two perspectives: probability distribution and an attention map, which represents intermediate information.
The proposed method consists of an ensemble mechanism for constructing an ensemble in a knowledge transfer graph to optimize the loss design between nodes.
Figure \ref{fig:1} shows an overview of the proposed method.
We explore how to learn to improve ensemble accuracy by optimizing a knowledge-transfer graph.

Our contributions are as follows:
\begin{itemize}
\item We carry out mutual learning for ensemble learning using loss designs that promotes diversity among networks. We evaluated several loss designs and confirmed that ensemble accuracy improves.
\item We propose an ensemble method using knowledge transfer that consists of an ensemble mechanism and a loss design in a knowledge-transfer graph to promote diversity and investigate various ensemble-learning methods.
The optimized knowledge-transfer graph was evaluated on the datasets of fine-grained object classification tasks, which confirmed that the proposed method results in be improved ensemble accuracy than the ensemble method using individually trained networks.
\end{itemize}

\section{Related work}  
\subsection{Ensemble method}
Ensemble method is one of the oldest machine-learning methods.
Ensemble method in deep learning is a simple method of averaging the probability distributions or logits output by networks with different weight parameters.
Ensemble accuracy improves as the number of networks included increases, and after a certain number of networks, ensemble accuracy stops improving.
Hyperparameter ensemble \cite{HypEns} shows that ensemble accuracy can be improved by combining the initial values of the network weights and hyperparameters. 
For ensemble method in Few-Shot learning \cite{FewShotEns}, in addition to applying different randomization to each network, we introduced a loss design in which the probability distributions of the networks are brought closer together and further apart and showed that bringing the networks closer together is more effective when the number of networks in an ensemble is small, and separating them is more effective when the number of networks in an ensemble is large. 
Due to the nature of ensembles that use multiple networks, they are also effective in various problem settings such as adversarial attack and out-of-distribution detection \cite{semi-supEns,AdversarialEns,OutOfDisRns}.

The learning and inference cost of ensembles increases with the number of networks. 
In knowledge transfer \cite{CL,ONE}, training a single network to mimic the ensembled probability distribution has been shown to perform as well as an ensemble with a single network. In batch ensemble \cite{BatchEns} and hyperparameter ensembles \cite{HypEns}, the number of parameters is prevented from increasing by commonizing some of the parameters, thereby reducing the training and inference costs.

\subsection{Knowledge transfer}
Knowledge transfer is a learning method in which a network shares the knowledge it has acquired through learning with the goal of network compression and network performance improvement.
There are two types of knowledge-transfer methods, i.e., unidirectional and bidirectional.

Unidirectional knowledge transfer uses a teacher network, which is a network that has been trained, and a student network, which is an untrained network.
In addition to the teacher label, the output of the teacher network is used as a pseudo-supervisor label to train the student network.
Hinton et al. \cite{KD} proposed knowledge distillation (KD), which uses the probability distribution of a teacher network with a large number of parameters to train a student network with a small number of parameters.
This method is also effective for teacher and student networks with the same number of parameters \cite{Born_Again_Net}. A two-stage knowledge-transfer method using three networks has also been proposed \cite{TA_distillation}.

Bidirectional knowledge transfer uses the probability distributions output by networks other than as pseudo-supervisory labels by using only the student networks. The first bidirectional knowledge-transfer method was proposed by Zhang et al. \cite{DML} called deep mutual learning (DML).

In addition to probability distributions, methods have been proposed that use intermediate layer features and relationships between samples as knowledge \cite{FitNets,IRG,FSP_matrix,similarity-preserving,correlation_congruence,VID,Learning_Metrics_from_Teachers,CRD,AT,Grad-CAM++}. Knowledge-transfer methods have been used in a variety of problem settings \cite{detection,structured_knowledge_distillation,Mean_Teacher}.

\subsection{Knowledge-transfer graph}
A knowledge-transfer graph \cite{KTG} is used to consider a variety of knowledge transfer methods by using hyperparameter search, where the network type and knowledge-transfer method are hyperparameters. 
A knowledge-transfer graph represents knowledge-transfer methods as a graph, which is a unified representation of conventional methods. 
The networks are represented as nodes, and knowledge transfers are represented as directed edges. 
The direction of the effective edge represents the direction of the knowledge transfer. 
The destination of the knowledge transfer is called the target node, and the source of the knowledge is called the source node. 
If we assume two nodes, one of which is a learned network, and the edge is in one direction, we express a KD. 
By defining two nodes as untrained networks and edges as bidirectional, one expresses DML.

A knowledge-transfer graph executes various knowledge transfers by applying one of four gate functions (through gate, cutoff gate, linear gate, and correct gate) to the loss of knowledge transfer.

The through gate passes the loss of each input sample as it is and is defined as
\begin{align}
	G_{s,t}^{Though}(a)=a.
\end{align}
The cutoff gate does not execute loss calculation and is defined as
\begin{align}
	G_{s,t}^{Cutoff}(a)=0.
\end{align}
The linear gate changes the weights linearly with training time and is defined as
\begin{align}
	G_{s,t}^{Linear}(a)=\frac{k}{k_{end}}a,
\end{align}
where $k$ is the number of the current iterations and $k_{end}$ is the total number of iterations at the end of the training The correct gate passes only the samples that the source node answered correctly and is defined as
\begin{align}
	G_{s,t}^{Correct}(a)=
	    \left\{
	        \begin{array}{ll}
                a & y_s = \hat{y}\\
                0 & y_s \neq \hat{y}
            \end{array}
        \right..
\end{align}
where $y_s$ is the output of the source node and $\hat{y}$ is a label.

The number of nodes and types of networks are defined in advance, and the gate functions of the networks and edges used as nodes are optimized by hyperparameter search to consider the various knowledge-transfer method and obtain optimal graph.

\section{Proposed Method}
The proposed ensemble method uses knowledge transfer. 
Since ensembles average probability distributions and logits, we believe that the more diversity there is in the outputs between networks, the better the effect. 
The proposed method consists of a loss design where the outputs are separated. 
We investigated the ensemble effects of separating and combining network knowledge. 
The number of possible combinations of these loss designs is enormous and difficult to design manually. 
Therefore, we extend ensemble learning with mutual learning to the optimization of a knowledge-transfer graph.

We add ensemble nodes that form ensembles using conventional knowledge-transfer graphs. 
For edge loss computation, a loss term is added using the output of the middle layer in addition to the output of the final layer of a network. 
By optimizing the accuracy of the ensemble nodes to maximize accuracy, a variety of ensemble-learning methods can be considered.

\subsection{Mutual learning for ensembles}
The output used for knowledge transfer can be divided into probability distributions and middle features. 
In this study, we used attention maps as middle features. 
To learn as a minimization problem, we used a different loss design for each output when bringing them closer together and when separating them. 
The target network was the knowledge transfer destination and the source network was the knowledge source.

When the probability distributions is brought closer together, kullback-leibler(KL)-divergence is used, and when it is farther separated, cosine similarity is used. The loss function using KL-divergence is defined as
\begin{align}
	KL(\bm{p}_s(\bm{x})\parallel \bm{p}_t(\bm{x})=\sum_{c=1}^{C}p_s^{c}(\bm{x})\log\frac{p_s^{c}(\bm{x})}{p_t^{c}(\bm{x})},
\end{align}
\begin{align}
    \label{quad:1}
	L_{p}=\frac{1}{N}\sum_{n=1}^{N}KL(\bm{p}_s(\bm{x}_n)\parallel \bm{p}_t(\bm{x}_n)).
\end{align}
where $C$ is the number of classes, $N$ is the number of samples, $x$ is the input sample, $p_s(x)$ is the probability distribution of the source network, and $p_t(x)$ is the probability distribution of the target network. 
The loss function using cosine similarity is defined as
\begin{align}
    \label{quad:2}
	L_{p}=\frac{1}{N}\sum_{n=1}^{N}\frac{\bm{p}_s(\bm{x}_n)}{\parallel \bm{p}_s(\bm{x}_n)\parallel_2}\cdot \frac{\bm{p}_t(\bm{x}_n)}{\parallel \bm{p}_t(\bm{x}_n)\parallel_2}.
\end{align}

The attention map responds strongly to regions that are effective in learning the input sample.
Since the size of the target object in the sample varies, the size of the strongly responsive region differs depending on the sample. 
Thus, there are cases in which the similarity is high even though the object is responding strongly to different parts of the object. 
Therefore, we crop the attention map. 
The source network is cropped to a square centered on the pixel with the highest attention, and the target network is cropped to the same position as the source network. 
The cropping is done in multiple sizes, and the average of the similarities for each size is used as the similarity of the attention map. 
When the attention map is brought closer together, mean squared error is used, and when it is farther separated, cosine similarity is used. 
The loss function using mean squared error is defined as
\begin{align}
    \label{quad:3}
	L_{map}=\frac{1}{NK}\sum_{n=1}^{N}\sum_{k=1}^{K}(\frac{\bm{Q}_s^{k}(\bm{x}_n)}{\parallel \bm{Q}_s^{k}(\bm{x}_n)\parallel_2}-\frac{\bm{Q}_t^{k}(\bm{x}_n)}{\parallel \bm{Q}_t^{k}(\bm{x}_n)\parallel_2})^2,
\end{align}
where $K$ is the number of crops, $Q_s$ is the attention map of the source network, and $Q_t$ is the attention map of the target network. 
The loss function using cosine similarity is defined as
\begin{align}
    \label{quad:4}
	L_{map}=\frac{1}{NK}\sum_{n=1}^{N}\sum_{k=1}^{K}\frac{\bm{Q}_s^{k}(\bm{x}_n)}{\parallel \bm{Q}_s^{k}(\bm{x}_n)\parallel_2}\cdot \frac{\bm{Q}_t^{k}(\bm{x}_n)}{\parallel \bm{Q}_t^{k}(\bm{x}_n)\parallel_2}.
\end{align}

\begin{figure}[t]
\centering
\includegraphics[width=8.3cm]{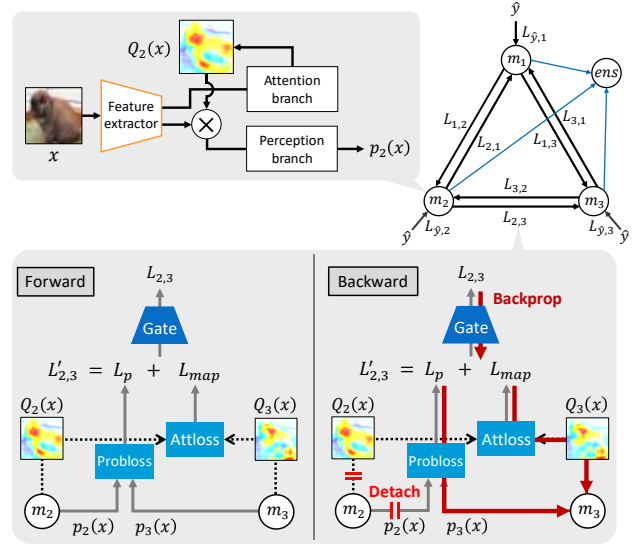}
\caption{Loss calculation in knowledge-transfer graph (for 3 nodes) with attention loss added. Loss calculation shows knowledge transfer from $m_2$ to $m_3$. Calculated loss gradient information is only propagated in $m_3$.}
	\label{fig:loss_f}
\end{figure}

\begin{figure*}[t]
    \begin{center}
        \begin{tabular}{c}
            \begin{minipage}{0.48\hsize}
                \begin{center}
                    \includegraphics[width=8cm]{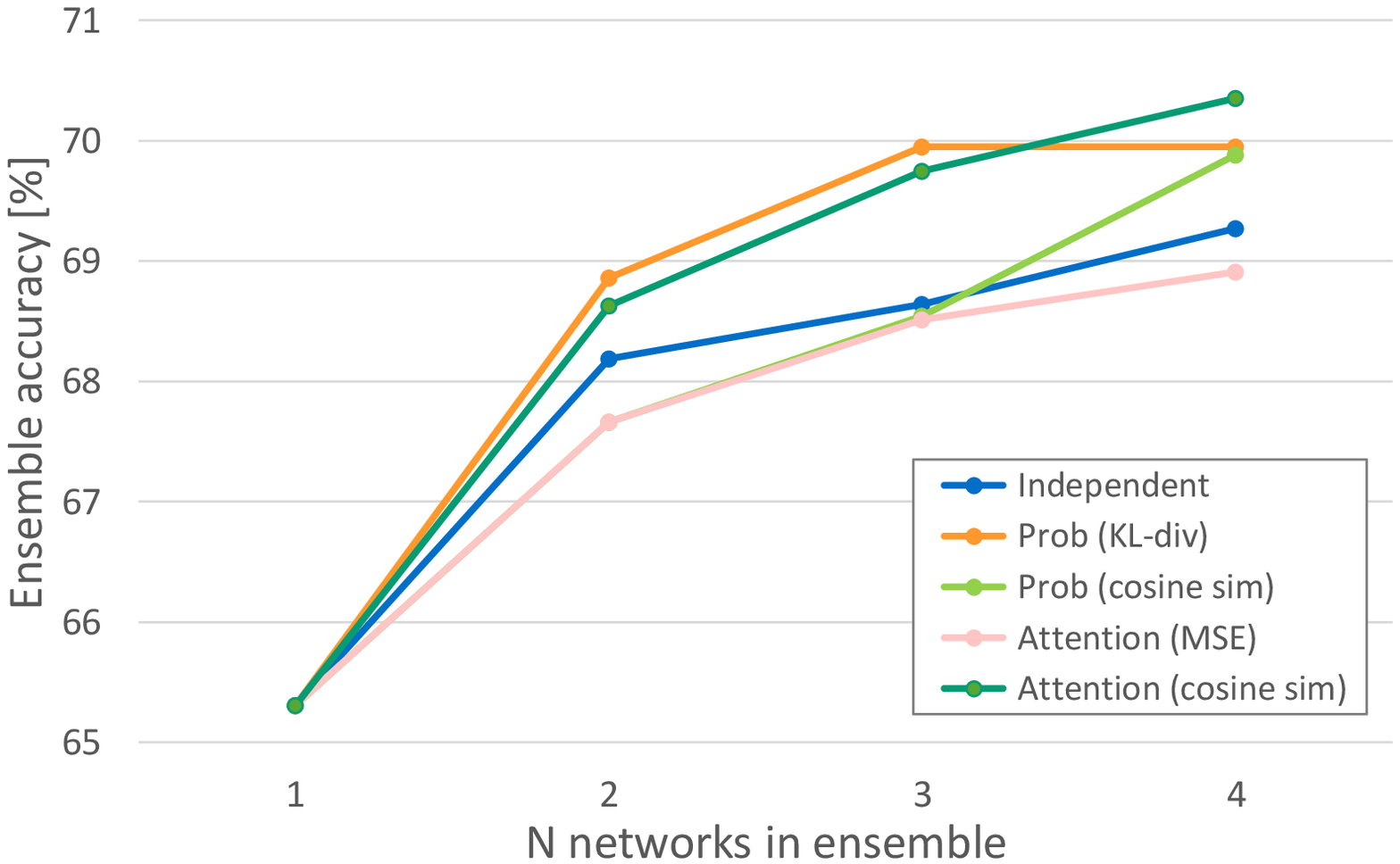}
                    \subcaption{ResNet-18}
                    \label{fig:ResNet}
                \end{center}
            \end{minipage}
            \begin{minipage}{0.48\hsize}
                \begin{center}
                    \includegraphics[width=8cm]{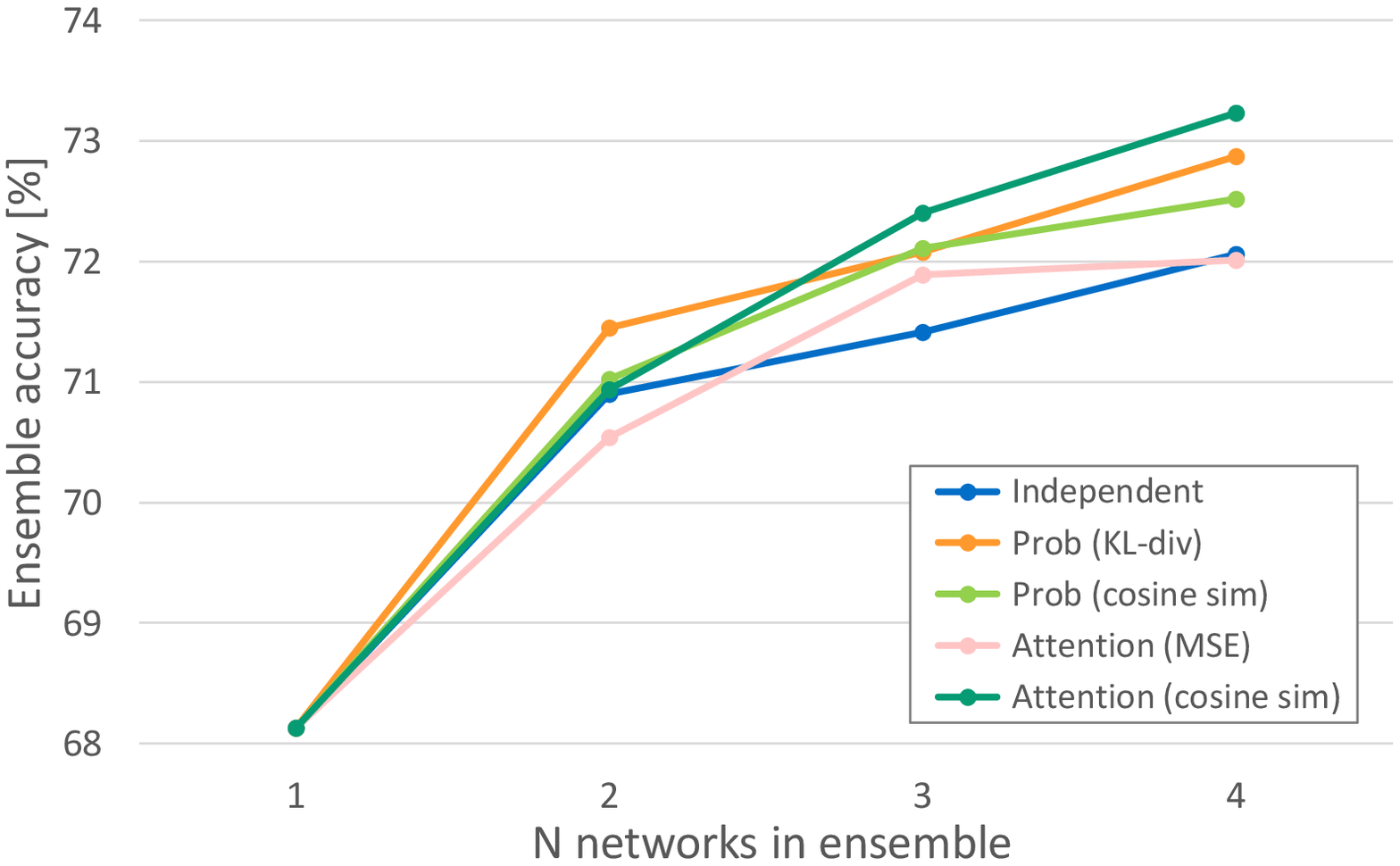}
                    \subcaption{ABN}
                    \label{fig:ABN}
                \end{center}
            \end{minipage}
        \end{tabular}
        \vspace{-2mm}
 \caption{Ensemble accuracy of loss design for various numbers of networks. Prob(KL-div) uses Eq. \ref{quad:1}, Prob(cosine sim) uses Eq. \ref{quad:2}, Attention(MSE) uses Eq. \ref{quad:3}, and Attention(cosine sim) uses Eq. \ref{quad:4}.}
        \label{fig:2}
        \vspace{-2mm}
    \end{center}
\end{figure*}

\subsection{Knowledge-transfer graphs for ensembles}
A knowledge-transfer graph selects the nodes for which we want to improve accuracy and executes a hyperparameter search. 
We add an ensemble node with a mechanism for constructing ensemble to the knowledge transfer graph to target the ensemble. 
The ensemble node forms an ensemble using the outputs of all the nodes. 
The ensemble node is defined as
\begin{align}
	\bm{y}_{ens}=\frac{1}{M}\sum_{m=1}^{M} \bm{y}_m(\bm{x}),
\end{align}
where $M$ is the number of nodes, $y_m$ is the logits output by the nodes, and $x$ is the input sample.

We add knowledge transfer using the attention map to the edge between nodes. 
Figure \ref{fig:loss_f} shows the process flow of loss calculation. 
In the conventional edge between nodes, the loss per sample is calculated using the probability distribution output by the node. 
Next, the gate function is applied to the loss. 
Knowledge transfer using the attention map is also carried out in the same manner. 
The loss for each sample is calculated using the attention map obtained from the node. 
Next, the gate function is applied to the losses. 
The gate function is the same for the loss of the probability distribution and that of the attention map. 
Finally, the loss of the probability distribution and that of the attention map are added. 
The loss function at the edge between the nodes is defined as
\begin{align}
	L_{s,t}=\frac{1}{N}\sum_{n}^{N}G_{s,t}(L_{p}(\bm{x}_n)+L_{map}(\bm{x}_n)),
\end{align}
where $G_{s,t}(\cdot)$ is the gate function, $L_{p}$ is the loss of the probability distribution, and $L_{map}$ is the loss of the attention map.
The loss is calculated for each edge. The final node loss is defined as
\begin{align}
	L_{t}=L_{hard}+\sum_{s=1,s\neq t}^{M}L_{s,t},
\end{align}
where $L_{hard}$ is the cross-entropy loss between the probability distribution output by the node and label.

The hyperparameters of a knowledge-transfer graph are the loss design of the probability distribution, that of the attention map, and the gate function. 
There are six combinations of losses between edges: bring the probability distribution closer to it of other edge (Eq. \ref{quad:1}), separate the probability distribution (Eq. \ref{quad:2}), bring the attention map closer to it of other edge (Eq. \ref{quad:3}), separate the attention map (Eq. \ref{quad:4}), Bring the probability distribution and attention map closer  to it of other edge at the same time (Eqs. \ref{quad:1} and \ref{quad:3}), and Separate the probability distribution and attention map at the same time (Eqs. \ref{quad:2} and \ref{quad:4}). The network to be used as a node is fixed to that determined before optimization.

Random search and the asynchronous successive halving algorithm (ASHA) are used to optimize a knowledge-transfer graph. 
The combination of hyperparameters is determined randomly, and the knowledge-transfer graph evaluates the ensemble nodes at $1, 2, 4, 8 \cdots 2k$ epochs. 
If the accuracy of the ensemble node is less than the average accuracy at the same epoch in the past, the learning is terminated and the next knowledge-transfer graph is trained.

\begin{figure*}[h]
    \begin{center}
        \begin{tabular}{c}
            \begin{minipage}{0.48\hsize}
                \vspace{7mm}
                \begin{center}
                	\includegraphics[width=6cm]{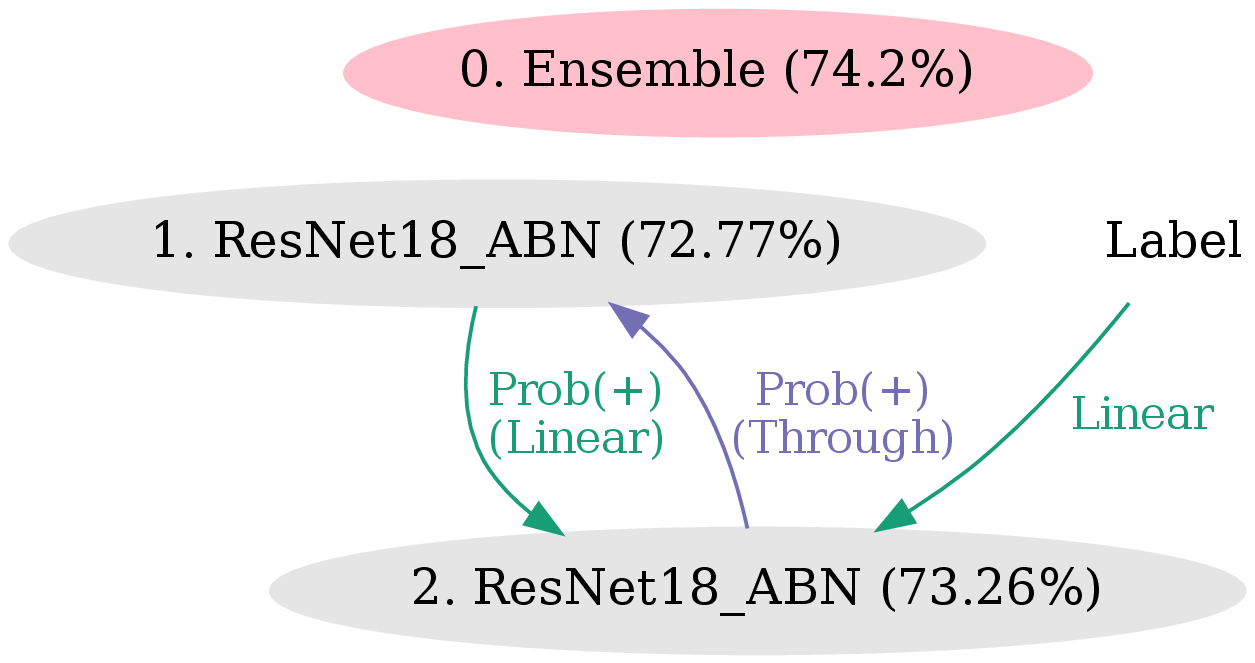}
                	\vspace{5mm}
 	\subcaption{2 nodes}
                    \label{fig:02models}
                \end{center}
            \end{minipage}
            \begin{minipage}{0.48\hsize}
                \begin{center}
                	\includegraphics[width=7cm]{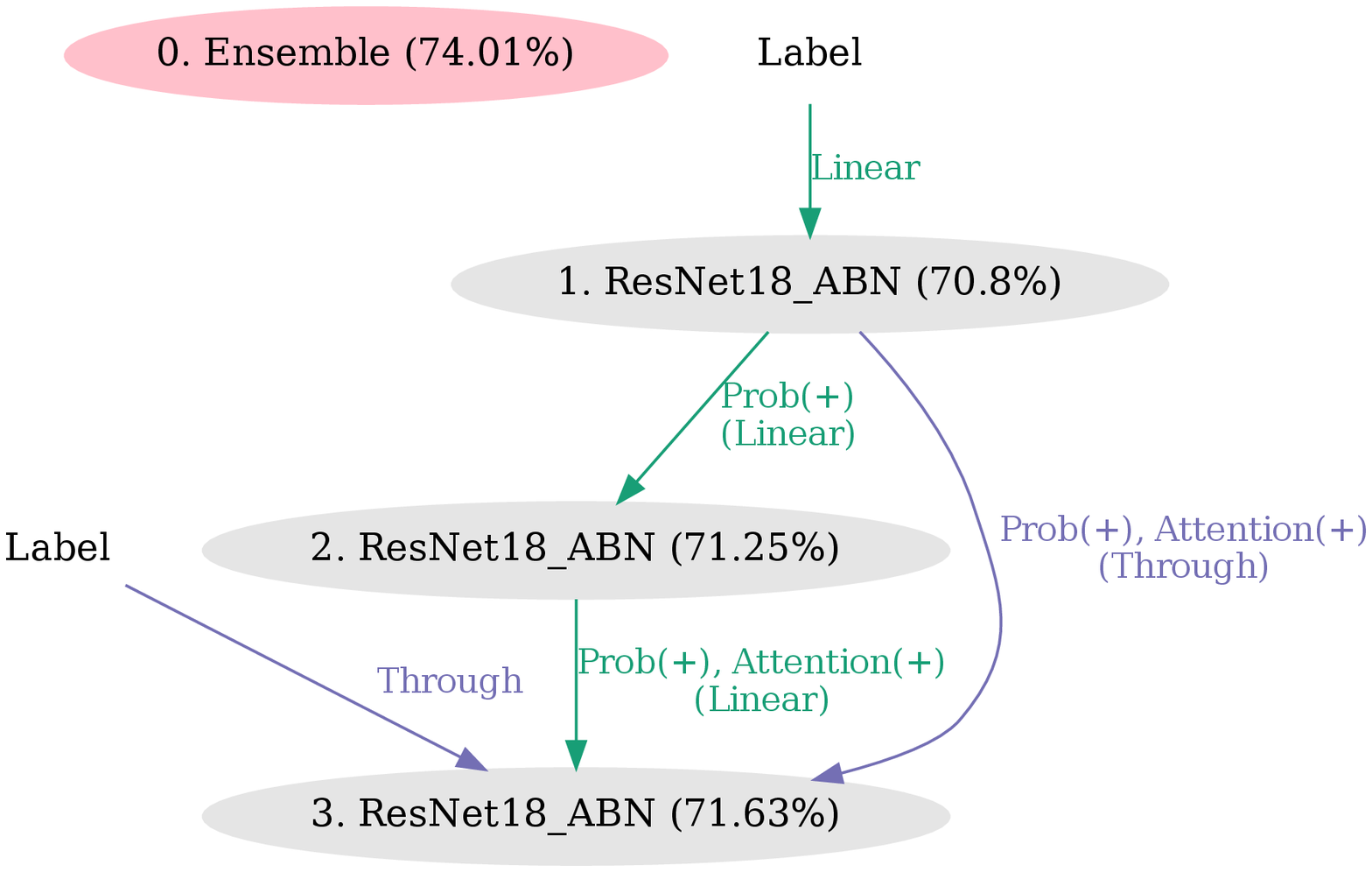}
                	\vspace{-1mm}
 	\subcaption{3 nodes}
                	\label{fig:03models}
                \end{center}
            \end{minipage}
            \\
            \begin{minipage}{0.48\hsize}
                \vspace{4mm}
                \begin{center}
                	\includegraphics[width=8cm]{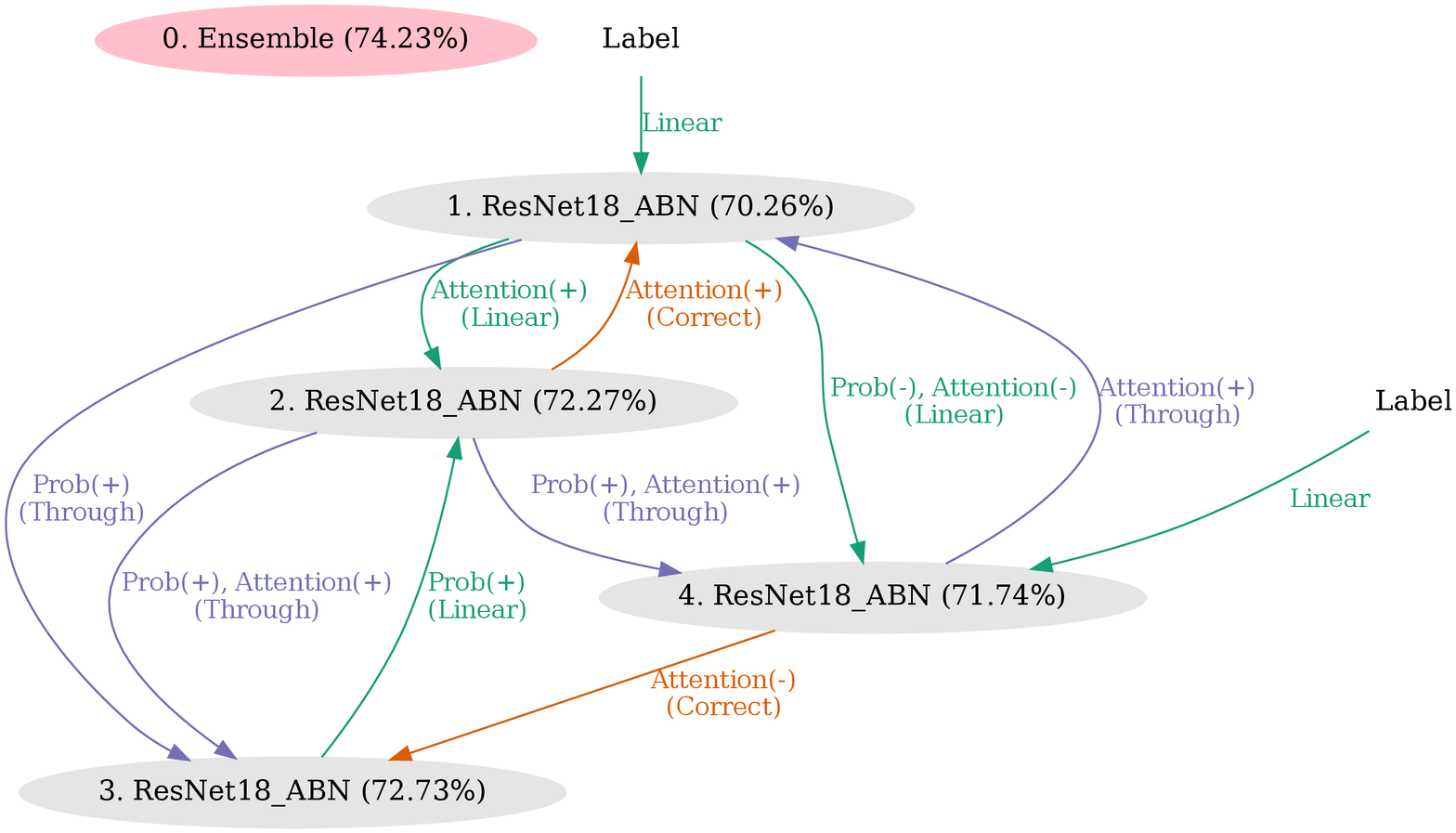}
 	\vspace{0 mm}
                	\subcaption{4 nodes}
                	\label{fig:04models}
                \end{center}
            \end{minipage}
            \begin{minipage}{0.52\hsize}
                \vspace{5mm}
                \begin{center}
                	\includegraphics[width=9cm]{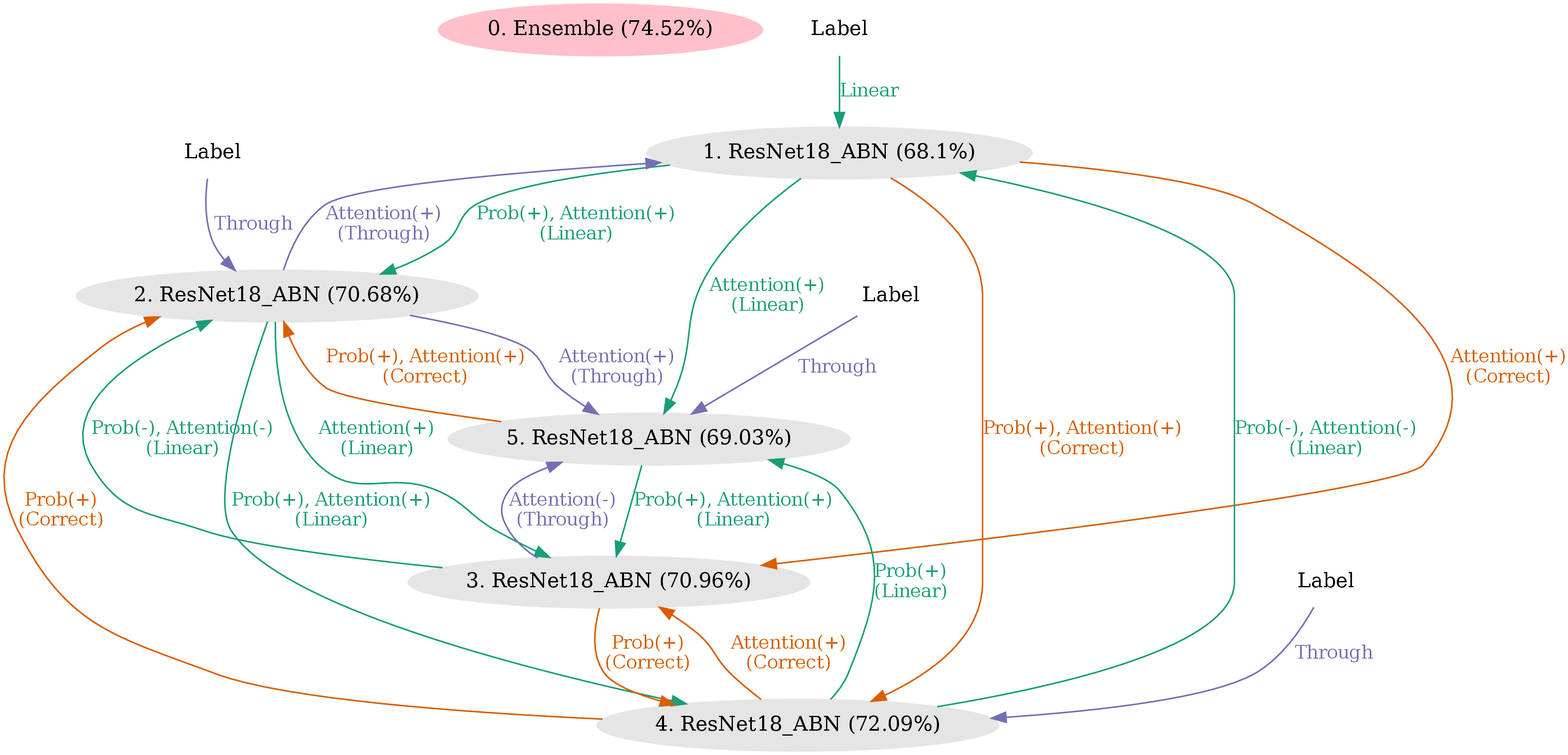}
                	\vspace{1mm}
 	\subcaption{5 nodes}
                	\label{fig:05models}
                \end{center}
            \end{minipage}
        \end{tabular}
        \vspace{-2mm}
\caption{Knowledge-transfer graph optimized on Stanford Dogs. Red node represents ensemble node and “Label” represents supervised labels. At each edge, selected loss design and gate are shown, exclusive of cutoff gate. Accuracy in parentheses is the result of one of five trials.}
        \label{fig:3}
    \end{center}
    \vspace{-2mm}
\end{figure*}

\section{Experiments}
We evaluated the proposed method.
Section 4.2 describes ensemble learning with mutual learning, Section 4.3 visualizes optimized knowledge transfer graphs, Section 4.4 compares the proposed method with conventional ensemble methods, and Section 4.5 evaluates the generalizability of knowledge-transfer graphs for different datasets.

\subsection{Experimental setting}
{\bf Datasets} \, The datasets we used were Stanford Dogs \cite{SDogs}, Stanford Cars \cite{SCars}, and CaltechUSCD Birds (CUB-200-2011) \cite{CUB2011}. 
These datasets belong to the fine-grained object classification task. 
Stanford Dogs consists of 120 classes of dog breeds and uses 12,000 images for training and 8,580 images for testing. 
Stanford Cars consists of 196 classes of cars and uses 8,144 images for training and 8,041 images for testing. 
CUB-200-2011 consists of 200 classes of birds and uses 5,994 images for training and 5,794 images for testing. 
To optimize a knowledge-transfer graph, half of the training data, selected to maintain class balance in the dataset, were used for training during the exploration of graph, and the rest of the data were used for evaluation during the exploration of graph. 
For the comparative evaluation discussed in Sections 4.4 and 4.5, the original training data and testing data were used.

{\bf Networks} \, For the networks, we used ResNet-18\cite{ResNet} and attention branch network (ABN) \cite{ABN} based on ResNet. 
We used attention transfer (AT) \cite{AT} to create an attention map in ResNet-18. 
AT is a knowledge-transfer method that creates an attention map by averaging feature maps in the channel direction. 
The attention map is created using the feature map output using the fourth ResBlock and crops it to sizes $3\times3$ and $5\times5$ for loss calculation. 
ABN creates an attention map by using a method based on the class activation map \cite{CAM} and weights the attention map to the feature map by using the attention mechanism. The attention map is cropped to $3\times3$, $7\times7$, and $11\times11$ for loss calculation. 
We believe that since the attention map is applied to the attention mechanism, the effect of the attention map on the probability distribution is better transmitted.
In the attention mechanism, residual processing is used to prevent loss of features due to the attention map, however, in the experiment, residual processing was not used to convey the changes in the attention map more strongly to the feature map.

{\bf Implementation details} \, The learning conditions were the same for all experiments. 
The optimization algorithms used were stochastic gradient descent and momentum. 
The initial learning rate was 0.1, momentum was 0.9, coefficient of weight decay was 0.0001, batch size was 16, and number of epochs was 300. 
The learning rate was decayed by a factor of 10 at 150 epochs and 225 epochs.
In the optimization of the knowledge-transfer graph, we tried 6,000 combinations of hyperparameters.
We used PyTorch \cite{pytorch} as a framework for deep learning and Optuna \cite{Optuna} as a framework for hyperparameter search. For the optimization of a knowledge-transition graph, we used 90 Quadro P5000 servers. 
Each result represents the mean and standard deviation of five trials.


\begin{figure*}[t]
    \begin{center}
        \begin{tabular}{c}
            \begin{minipage}{0.48\hsize}
                \begin{center}
                    \includegraphics[width=8.2cm]{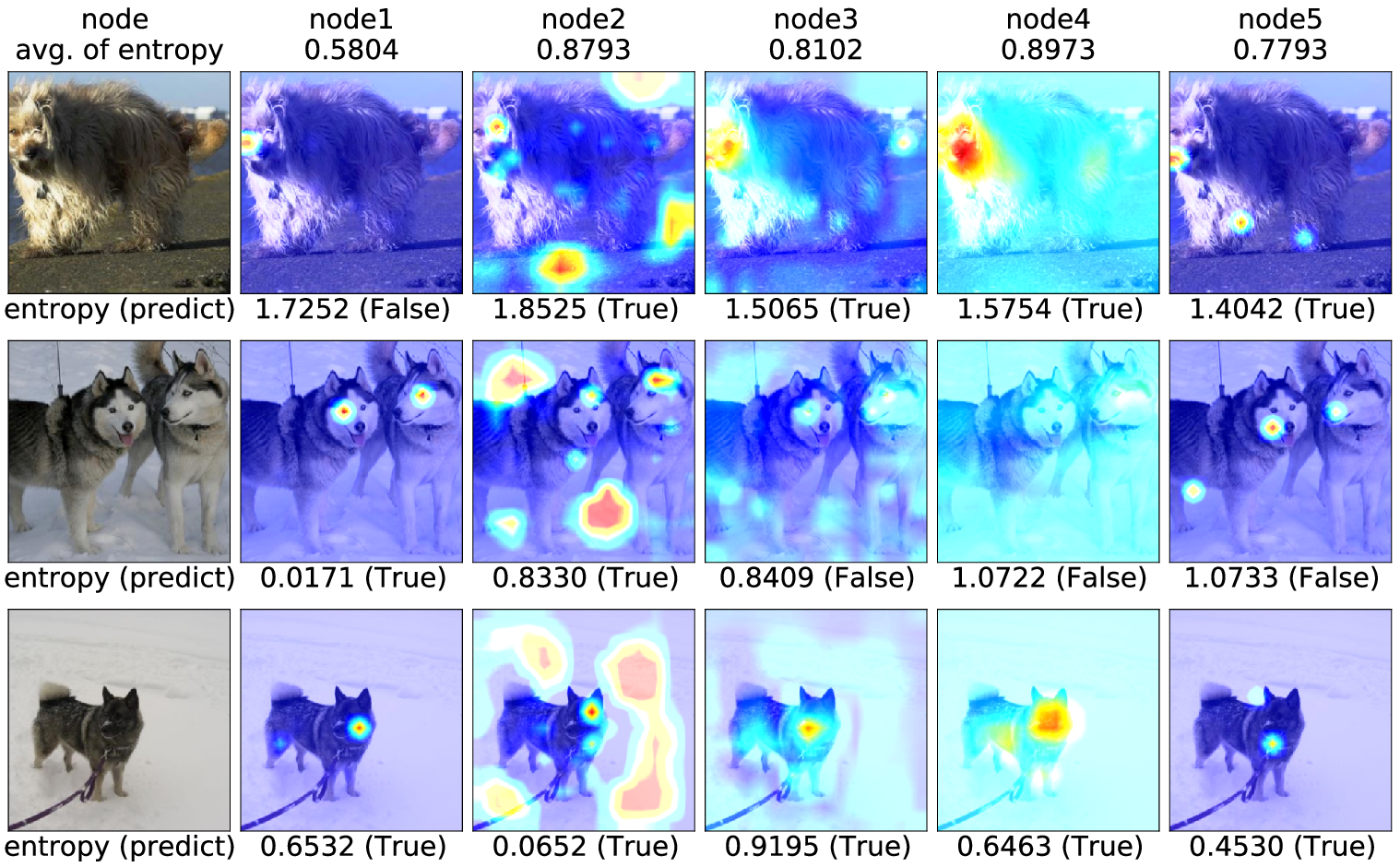}
                    \subcaption{Ours}
                    \label{fig:att_our}
                \end{center}
            \end{minipage}
            \begin{minipage}{0.48\hsize}
                \begin{center}
                    \includegraphics[width=8.2cm]{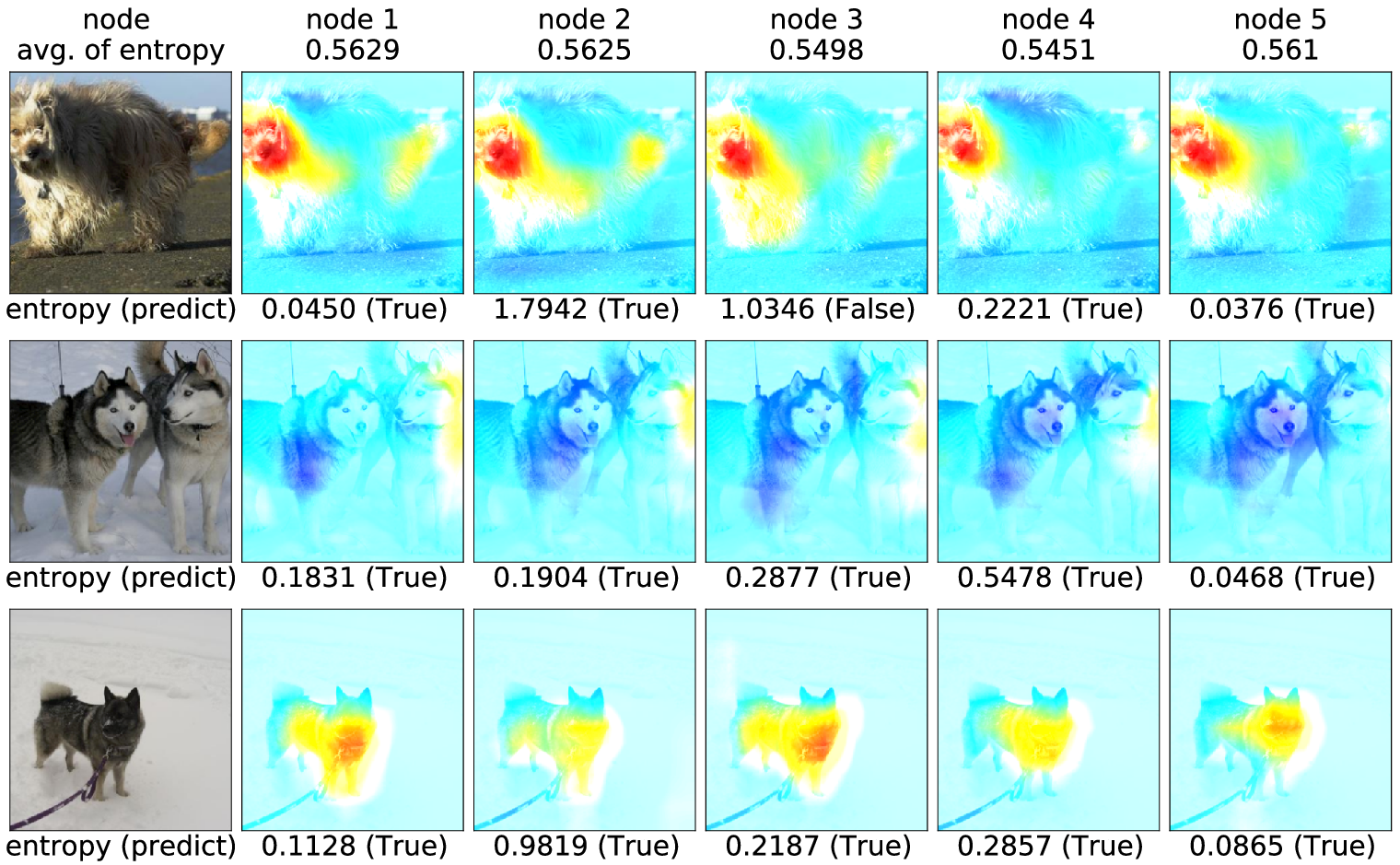}
                    \subcaption{Independent}
                    \label{fig:att_indep}
                \end{center}
            \end{minipage}
        \end{tabular}
        \vspace{-2mm}
 \caption{Attention map of optimized knowledge-transfer graph with 5 nodes and the ensemble method using individually trained networks. Bottom of map shows prediction results and entropy of probability distribution.}
        \label{fig:attention}
        \vspace{-2mm}
    \end{center}
\end{figure*}

\subsection{Ensemble learning with mutual learning}
In addition to the cross-entropy loss with labels when training networks, one of Eqs.\ref{quad:1} to \ref{quad:4} is applied as a loss design between networks. 
The results using ResNet-18 are shown in Fig. \ref{fig:ResNet}. 
The results using ABN are shown in Fig. \ref{fig:ABN}. 
When the number of networks used in the ensemble was small, it was effective to bring the probability distribution closer together, and when the number of networks was large, it was effective to separate the probability distribution and the attention map. 
Therefore, it is effective to add diversity to the probability distribution and attention map in the ensemble.
For AT, the accuracy of the loss design was reversed at the four networks. 
For ABN, the accuracy of the loss design that separated the attention map exceeded that of the loss design that brought the probability distributions closer together at the three networks. 
Therefore, the change in the attention map has more impact on the probability distribution by weighting it to the feature map by using the attention mechanism. 

\begin{table*}[t]
\begin{center}
\caption{Comparison of proposed method with conventional methods on Stanford Dogs}
\vspace{-2mm}
\label{tab:acc}
\scalebox{0.95}[0.95]{
\begin{tabular}{c|c|c|c|c|c}
\hline
\multirow{2}{*}{~Method~} & \multirow{2}{*}{Node} & \multirow{2}{*}{Node-to-node loss design} & \multirow{2}{*}{Gate} & \multicolumn{2}{c}{Accuracy {[}\%{]}} \\ \cline{5-6} 
& & & & \multicolumn{1}{c|}{Average of nodes} & ~Ensemble~ \\ \hline \hline
Independent & ResNet18 $\times$ 2 & - & - & 65.31~$\pm$~0.16 & 68.19~$\pm$~0.20 \\
Independent & ABN $\times$ 2 & - & - & 68.13~$\pm$~0.16 & 70.90~$\pm$~0.19 \\
DML & ABN $\times$ 2 & Prob(KL-divergence) & Fixed(Through) & 69.91~$\pm$~0.46 & 71.45~$\pm$~0.52 \\
Ours & ABN $\times$ 2 & Optimized & Optimized & \bf{72.77~$\pm$~0.23} & \bf{73.86~$\pm$~0.26} \\ \hline
Independent & ResNet18 $\times$ 3 & - & - & 65.08~$\pm$~0.23 & 68.64~$\pm$~0.38 \\
Independent & ABN $\times$ 3 & - & - & 68.04~$\pm$~0.28 & 71.41~$\pm$~0.34 \\
DML & ABN $\times$ 3 & Prob(KL-divergence) & Fixed(Through) & 70.50~$\pm$~0.26 & 72.08~$\pm$~0.42 \\
Ours & ABN $\times$ 3 & Optimized & Optimized & \bf{70.95~$\pm$~0.16} & \bf{73.41~$\pm$~0.30} \\ \hline
Independent & ResNet18 $\times$ 4 & - & - & 65.29~$\pm$~0.35 & 69.27~$\pm$~0.49 \\
Independent & ABN $\times$ 4 & - & - & 68.30~$\pm$~0.27 & 72.06~$\pm$~0.53 \\
DML & ABN $\times$ 4 & Prob(KL-divergence) & Fixed(Through) & \bf{71.50~$\pm$~0.31} & 72.87~$\pm$~0.29 \\
Ours & ABN $\times$ 4 & Optimized & Optimized & 71.46~$\pm$~0.22 & \bf{74.16~$\pm$~0.22} \\ \hline
Independent & ResNet18 $\times$ 5 & - & - & 65.00~$\pm$~0.24 & 69.47~$\pm$~0.13 \\
Independent & ABN $\times$ 5 & - & - & 68.24~$\pm$~0.26 & 72.32~$\pm$~0.18 \\
DML & ABN $\times$ 5 & Prob(KL-divergence) & Fixed(Through) & \bf{71.15~$\pm$~0.28} 
& 72.50~$\pm$~0.16 \\
Ours & ABN $\times$ 5 & Optimized & Optimized & 70.23~$\pm$~0.33 & \bf{74.14~$\pm$~0.50} \\ \hline
\end{tabular}
}
\end{center}
\end{table*}

\begin{table*}[]
\begin{center}
\caption{Ensemble accuracy of reused graphs optimized on another dataset. Graphs were trained on CUB-200-2011 and Stanford Cars, where graphs were explored on Stanford Dogs or datasets used for training.}
\label{tab:acc2}
\vspace{-1.5mm}
\begin{tabular}{c|c|cccc}
\hline 
Training & Exploring & \multicolumn{4}{c}{Number of nodes} \\ \cline{3-6} 
graph & graph & 2 & 3 & 4 & 5 \\ \hline \hline
CUB-200-2011  & Stanford Dogs & 72.06~$\pm$~0.16 & 71.82~$\pm$~0.25 & 73.03~$\pm$~0.11 & 72.13~$\pm$~0.17 \\
CUB-200-2011  & CUB-200-2011  & 69.81~$\pm$~0.38 & 74.17~$\pm$~0.23 & 72.22~$\pm$~0.30 & 74.05~$\pm$~0.58 \\ \hline
Stanford Cars & Stanford Dogs & 89.76~$\pm$~0.25 & 89.94~$\pm$~0.19 & 89.98~$\pm$~0.13 & 90.41~$\pm$~0.11 \\
Stanford Cars & Stanford Cars & 89.44~$\pm$~0.09 & 90.04~$\pm$~0.09 & 89.57~$\pm$~0.19 & 90.73~$\pm$~0.19 \\ \hline
\end{tabular}
\end{center}
\vspace{-3mm}
\end{table*}

\subsection{Visualization of optimized knowledge-transfer graphs}
Figure \ref{fig:3} shows the knowledge-transfer graphs of 2 to 5 nodes optimized on Stanford Dogs. 
At node 2, we obtained a graph that is an extension of DML.
At node 3, we obtained a graph that combines the conventional knowledge-transfer methods of KD and TA.
For 4 and 5 nodes, we obtained graphs with a mixture of loss designs that are brought closer together and loss designs that are separated.

Figure \ref{fig:att_our} shows the attention map in the optimized knowledge-transfer graph with five nodes.
Each node has a different focus of attention.

Looking at the average entropy, nodes 1 and 5, which focus on a single point on the dog’s head, have low entropy.
Nodes 2, 3, and 4, which focus on the whole image or background, have higher entropy than nodes 1 and 5.
This means that inferences are made on the basis of the importance of different locations and the state of attention affects probability distribution.

Figure \ref{fig:att_indep} shows the attention map in the ensemble method using individually trained networks.
Comparred to the optimized knowledge-transfer graph, the average entropy of the ensemble method using individually trained networks is lower.
This is because the attention regions are almost the same among the networks even though they are trained individually.

\subsection{Comparison with conventional methods}
Table \ref{tab:acc} shows the average and ensemble accuracies of the nodes of the proposed and conventional methods for Stanford Dogs. 
"Ours" is the result of the optimized knowledge-transfer graph with the proposed method, "Independent" is the result of the individually trained network, and "DML" is the result of the network with DML. 
The ensemble accuracy of "Ours" was higher than that of "Independent" and "DML" at any number of nodes. 
Comparing "Independent(ABN)" and "DML", we can see that the improvement in ensemble accuracy was small compared with the improvement in node accuracy.
With "Ours", compared with "DML", ensemble accuracy also improved as network accuracy improved.
Therefore, we can say that "Ours" obtained the graph that generates more diversity by learning.

\begin{figure}[t]
	\centering
	\includegraphics[width=8.3cm]{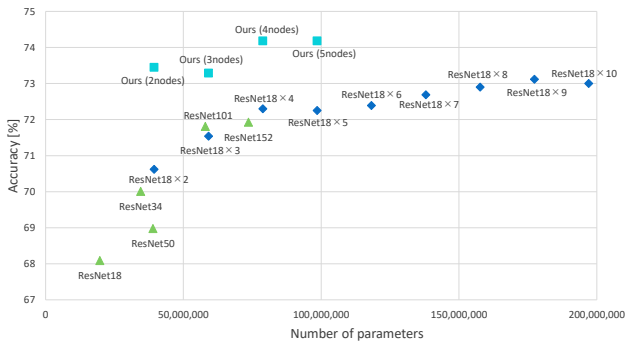}
	\caption{Relationship between number of parameters and accuracy in Stanford Dogs. Green shows single network, blue shows "Independent", and light blue shows "Ours".}
	\label{fig:4}
\end{figure}

Figure \ref{fig:4} shows the comparison results with ABN and different base networks. 
The vertical axis is accuracy and the horizontal axis is the total number of parameters. 
In Stanford Dogs, the accuracy of the single network and "Independent" varied with the number of parameters. 
The knowledge-transfer graph shows that the ensemble with high parameter efficiency can be constructed by mutual learning with diversity without changing the network structure. 
When the number of networks is increased, ensemble accuracy reaches a ceiling of around 73\%. 
This shows that the proposed method achieved an accuracy that exceeds the limit with a conventional method.

\subsection{Generalizability of graphs}
We evaluated the optimized graph structure using Stanford Dogs on various datasets. 
Table \ref{tab:acc2} shows the ensemble accuracies for CUB-200-2011 and Stanford Cars. 
The accuracies of the knowledge-transfer graphs with different datasets used for optimization are comparable. 
This shows that the structure of the optimized graph is generalizable. 
The accuracies of some knowledge-transfer graphs differ depending on the data used in the optimization. 
We believe this is caused by a large number of combinations of graph structures, which eventually results in the acquisition of different graph structures.

\begin{figure}[t]
	\centering
	\includegraphics[width=8cm]{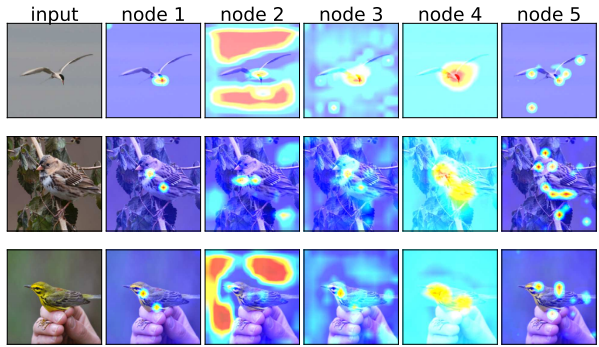}
	\vspace{-1.5mm}
	\caption{Attention map on CUB-200-2011 of graph with 5 nodes optimized by Stanford Dogs}
	\label{fig:CUB_att}
	\vspace{-1mm}
\end{figure}

\begin{figure}[t]
	\centering
	\includegraphics[width=8cm]{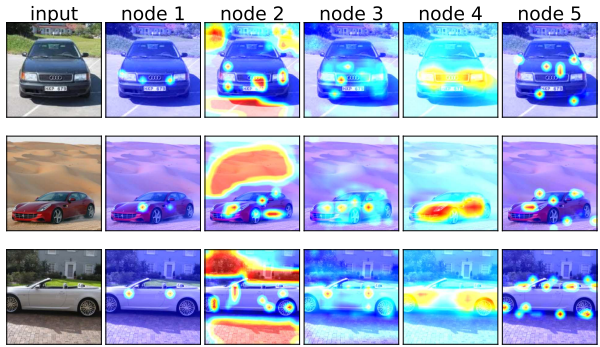}
	\vspace{-1.5mm}
	\caption{Attention map on Stanford Cars of graph with 5 nodes optimized by Stanford Dogs}
	\label{fig:SCars_att}
	\vspace{-2mm}
\end{figure}

Figures \ref{fig:CUB_att} and \ref{fig:SCars_att} shows the attention maps of the knowledge-transfer graph with five nodes optimized by Stanford Dogs. 
The attention maps show a similar trend to that in Figure \ref{fig:att_our} when trained on a different dataset than the exploration. 
This shows the generalizability of the structure of the optimized graph in terms of the attention map that is eventually obtained.

\section{Conclusion and Future Work}
We proposed knowledge-transfer graph for ensemble method.
We added the loss design as a hyperparameter to promote diversity among networks and optimized the graph structure to improve ensemble accuracy. 
We evaluated our method on a dataset of a fine-grained object classification task and the ensemble accuracy of our method was higher than that the result of the network with DML and the result of the individually trained network. 
For the number of hyperparameter combinations, 6,000 pairs were selected by random search and the branches were pruned using the ASHA. 
Since the number of combinations of graph structures increases in proportion to the number of nodes, learning methods with better performance may be obtained by increasing the number of trials. 
In the future, we will study the introduction of Bayesian optimization and its application to different tasks.


{\small
\bibliographystyle{ieee_fullname}
\bibliography{main}
}

\end{document}